\documentclass{article}

\PassOptionsToPackage{numbers, compress}{natbib}
\usepackage[preprint]{neurips_2026}

\usepackage[utf8]{inputenc} 
\usepackage[T1]{fontenc}    
\usepackage{hyperref}       
\usepackage{url}            
\usepackage{booktabs}       
\usepackage{amsfonts}       
\usepackage{nicefrac}       
\usepackage{microtype}      
\usepackage{xcolor}         

\usepackage{amsmath} 
\usepackage[bb=dsserif]{mathalpha} 
\usepackage{lipsum}
\usepackage{appendix}
\usepackage{cleveref}
\usepackage{xspace} 

\usepackage{graphicx}

\newcommand{\cmt}{{\footnotesize\textsc{CMT }\xspace}}
\newcommand{\fq}{{\footnotesize\textsc{FinQA }\xspace}}
\newcommand{\claude}{{\footnotesize\textsc{Claude Opus 4.5 }\xspace}}
\newcommand{\gpt}{{\footnotesize\textsc{GPT 5.5 }\xspace}}
\newcommand{\cfa}{{\footnotesize\textsc{CFA Level 1 }\xspace}}
\newcommand{\cfat}{{\footnotesize\textsc{CFA Level 2 }\xspace}}

\title{When Judges Agree and Are Wrong: Consistency-Accuracy Dissociation as a Blind Spot for Label-Free Evaluation}

\author{
    Xiang Zhang \\
   The Vanguard Group, Inc. \thanks{© 2026 The Vanguard Group, Inc. All rights reserved. \\This material is provided for informational purposes only and is not intended to be investment advice or a recommendation to take any particular investment action.}\\
  \texttt{xiang\_zhang@vanguard.com} \\ 
  \And
  Lovedeep Gondara \\
   The Vanguard Group, Inc. \\
  \texttt{lovedeep\_gondara@vanguard.com} \\
  \AND
  Jekaterina Novikova \\
   The Vanguard Group, Inc.\\
  \texttt{jekaterina\_novikova@vanguard.com} \\
}

\begin{document}

\maketitle


\begin{abstract}
Label-free evaluation methods, such as self-consistency filtering, majority-vote aggregation, and agreement-based reliability scoring, rely on an implicit assumption that consistency can serve as a proxy for correctness when ground-truth labels are unavailable or expensive to obtain. In this paper, we test this assumption directly. Using four financial QA datasets (multiple-choice and free-form), ten consistency dimensions, and two frontier model families, we measure the joint distribution of consistency and accuracy at the level of individual examples rather than in aggregate. We find that 60\% of all incorrect answers fall in the high-consistency region (consistency $\geq 0.8$), and that 77.7\% of these show complete answer collapse, i.e.\ identical wrong answers across all perturbation variants. This high-consistency/incorrect quadrant cannot be detected by any evaluation signal that is built on agreement alone, as the model appears maximally reliable while being wrong. We show that the phenomenon is robust to the choice of consistency threshold $t$ (it persists from $t=0.50$ to $t=1.00$), is shared across model families, and is concentrated in specific intersections of perturbation dimension and domain. We map where this quadrant occurs, analyze its stability properties, and discuss the implications for evaluation pipelines that use agreement-based signals in place of human labels.
\end{abstract}

\section{Introduction}
\vspace{-0.5em}
A growing family of evaluation methods treats consistency as a load-bearing proxy for correctness~\cite{Taubenfeld_2025,kadavath2022language,wang2023selfconsistencyimproveschainthought}. Self-consistency sampling, majority-vote filtering, and agreement-based reliability scores are deployed precisely in those settings where ground-truth labels are unavailable, and the underlying assumption is that a model which agrees with itself across meaning-preserving perturbations is more likely to be right. This assumption is rarely tested against ground truth~\cite{ding2026llms} at the resolution where it matters most: not on average across a benchmark, but conditionally, within the population of wrong answers. In this paper, we test it at that resolution and find that the assumption fails in the least desirable direction, as the majority of model errors are concentrated inside the high-consistency region that agreement-based methods would certify as reliable.

We situate this test in financial question answering, a domain in which the construct-validity question is both answerable and consequential. It is answerable because professionally validated exam items (CFA\footnote{https://www.cfainstitute.org/programs/cfa-program}, CMT\footnote{https://cmtassociation.org/}) provide unambiguous gold labels against which the dissociation can be measured. It is consequential because financial AI deployments are precisely the high-stakes, regulated settings~\cite{nedunuri2026paradox} in which consistency-based evaluation carries real operational weight: proprietary data constraints and compliance boundaries frequently prevent ground-truth labels from being shared~\cite{kulatilleke2022challenges, 10.1145/3786484.3786522}, which makes agreement-based proxies load-bearing rather than supplementary. The financial domain additionally provides internal variation that allows us to isolate what drives the dissociation: a difficulty gradient (\cfa\footnote{https://www.cfainstitute.org/programs/cfa-program/candidate-resources/level-i-exam} vs. \cfat\footnote{https://www.cfainstitute.org/programs/cfa-program/candidate-resources/level-ii-exam}), format differences (multiple-choice vs.\ free-form numerical reasoning), and distinct subject areas within a single professional knowledge domain. As such, we are able to attribute the concentration of the phenomenon to question complexity and consistency dimension rather than to confounded differences across unrelated benchmarks.

\begin{figure}[t]
    \centering
    \includegraphics[width=0.5\linewidth]{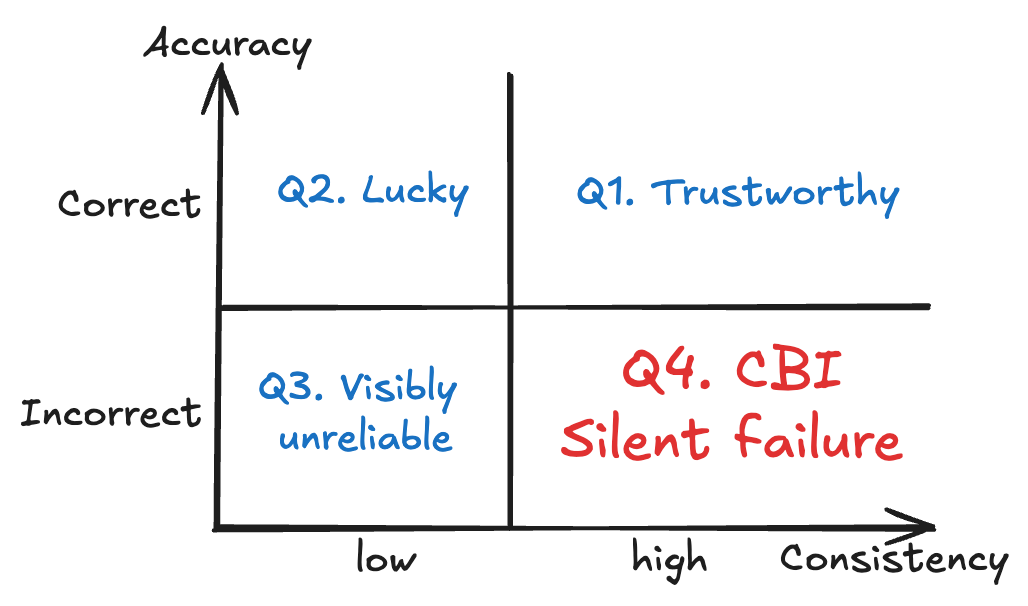}
    \caption{The joint consistency--accuracy space forms four quadrants: Q1 (high consistency, correct) corresponds to reliable behavior; Q2 (low consistency, correct) corresponds to accurate but inconsistent behavior; Q3 (low consistency, incorrect) corresponds to behavior that is unreliable but detectable; and Q4 (high consistency, incorrect) corresponds to CBI (Consistent But Inaccurate)}
    \label{fig:quadrant}
\end{figure}

We conduct our experiments across 4 financial QA datasets (\cfa and \cfat~\cite{stanishevskii2026finessebench}, \cmt~\cite{stanishevskii2026finessebench}, and \fq~\cite{chen2021finqa}), 10 consistency dimensions, 2 tested model families (\gpt and \claude), and 6 distinct generator-validator pipeline configurations, measuring the joint consistency--accuracy distribution at the level of individual examples. We treat the high-consistency/incorrect quadrant (Figure~\ref{fig:quadrant}) of this distribution as the primary object of study.

The contributions of this work are as follows. (1) We provide the comprehensive example-level \textbf{characterization of the dissociation between consistency and accuracy in financial QA}, and show that it is large and asymmetric: almost two-thirds (60\%) of all incorrect answers meet or exceed a consistency threshold of 0.80, which is the standard operating threshold for agreement-based filters, and 77.7\% of these Consistent But Inaccurate (CBI) examples show complete answer collapse, producing identical wrong responses across all perturbation variants. (2) We demonstrate that \textbf{the phenomenon is structural} rather than an artifact of experimental choices: it is threshold-invariant, with CBI rates moving only from 7.2\% to 5.05\% as the threshold increases from 0.50 to 1.00; it is shared across model families, which converge on the same wrong answers with consistency of 1.0 for specific questions and thereby defeat ensemble-based and cross-model agreement strategies; and it is stable across all 6 generator--validator configurations. (3) We \textbf{map where the phenomenon concentrates}, showing that CBI rates increase with question complexity (5.22\% for \cfat compared to 3.80\% for \cfa) and with answer format (7.49\% for free-form numerical reasoning compared to 5.42\% for multiple choice), while remaining diffuse across perturbation types, with rates between 0.2\% and 9.7\% in nine of ten dimensions and no single dimension dominating. (4) We \textbf{quantify the implications for evaluation pipelines} that substitute agreement-based signals for human labels, and derive from our analysis two practical recommendations: a low-cost pre-deployment diagnostic for estimating quadrant mass, and a structured disclosure format for reporting the failure modes of agreement-based evaluators. (5) Finally, we establish the \textbf{direct practical consequence} of the above findings: consistency-based filtering measurably improves the retained set, but any pipeline that uses it to certify data quality, to gate deployment decisions, or to substitute for human evaluation retains a majority of its errors inside the certified set, and those retained errors are disproportionately the deterministic ones that further sampling or ensembling cannot remove. The residual is therefore not incidental noise around a working signal but a stable and largely irreducible component of it.
\vspace{-0.5em}

\section{Related Work}
\vspace{-0.5em}
\citet{wang2023selfconsistencyimproveschainthought} introduce self-consistency, replacing greedy decoding with sampling of multiple reasoning paths and returning the plurality answer, on the premise that complex problems admit several routes to a single correct answer while errors diverge. \citet{Taubenfeld_2025} propose CISC, which weights sampled reasoning paths by self-assessed confidence rather than counting votes equally, matching or beating standard self-consistency while reducing the required number of paths. A parallel line of work measures consistency not over samples but over inputs: \citet{jang2022becel} decompose behavioral consistency into negational, symmetric, additive, and semantic sub-categories, providing the typology our perturbation dimensions follow. \citet{tan-etal-2025-consistent} identify self-consistent errors, where a model repeatedly returns the same wrong answer, and show that consistency-based detectors collapse to worse than chance on them. \citet{ding2026llms} audit self-consistency and cross-model agreement as confidence signals and conclude that agreement is a conditional proxy for correctness, with confident errors recurring across providers. \citet{norman2026reliability} conduct a large judge meta-evaluation and report kappa deflation: exact-match agreement overstates chance-corrected discrimination. \citet{krumdick2026freelabelslimitationsllmasajudge} introduce BFF-Bench, 160 business and finance questions authored by financial professionals, and VERDICTS, expert correctness annotations over 1,200 model responses. Their central result is that without a reference answer, judges agree with human experts primarily on questions the judges could answer correctly themselves. This work establishes that consistent errors exist and that agreement overstates validity; what remains unmeasured is the fraction of the \emph{error population} lying inside the certified region, its stability under threshold choice, and its concentration by perturbation type and difficulty.
\vspace{-0.5em}

\section{Method}
\vspace{-0.5em}
\noindent\textbf{Datasets.} We scope our experiments to financial QA for three reasons. First, it is a domain in which consistency-based evaluation is frequently used, since regulatory and data-sharing constraints frequently preclude access to ground-truth labels and make agreement-based proxies necessary. Second, professionally validated exam items (CFA, CMT) provide unambiguous gold labels against which the dissociation can be measured; this is a requirement our design depends on, and one that many specialized domains cannot satisfy. Third, the available datasets vary in difficulty (Level 1 vs.\ Level 2) and format (multiple-choice vs.\ free-form numerical reasoning), which allows us to analyze the results through these lenses. We use four financial QA datasets: \cfa and \cfat contain multiple-choice questions from the Chartered Financial Analyst curriculum, with Level 2 requiring multi-step quantitative reasoning; \cmt contains Chartered Market Technician exam items covering technical analysis and market structure; and \fq contains free-form numerical reasoning questions that require extraction and calculation over financial tables. Both \cfa, \cfat and \cmt datasets are accessed through the FINESSE-Bench benchmark~\cite{stanishevskii2026finessebench}. We use subsets of randomly selected 200 examples from each dataset, resulting in 23,170 individual example-level observations after perturbation expansion, which we explain next.

\noindent\textbf{Consistency dimensions.} Following the broader taxonomy of consistency phenomena in language models~\citep{novikova2025consistency}, each base item is perturbed along up to 10 consistency dimensions, each producing 3 variants: \textit{option order} (random permutation of answer labels)~\cite{zheng2024large,pezeshkpour-hruschka-2024-large}, \textit{option relabel} (substitution of label identifiers, e.g.\ A/B/C $\to$ X/Y/Z)~\cite{yang-etal-2025-option,zhou-etal-2024-revisiting}, \textit{paraphrase} (meaning-preserving restatement of the stem)~\citep{elazar2021measuring,choi2025roparq}, \textit{factual} (anchoring with additional true context)~\citep{xie2023adaptive}, \textit{additive} (insertion of irrelevant but plausible distractor text)~\citep{shi2023large}, \textit{crosslingual} (translation to Spanish, French, or German)~\citep{qi2023crosslingual}, \textit{symmetric} (swapping symmetric conditions in the problem)~\citep{asai2020logic}, \textit{negation} (semantic inversion of the question, requiring a different correct answer)~\citep{kassner2020negated}, \textit{distractor} (rewording of incorrect options)~\citep{zhang2024multiple,elhady2025wicked}, and \textit{self-consistency} (repeated presentation with varied reasoning prompts)~\cite{wang2023selfconsistencyimproveschainthought,kadavath2022language}. Perturbations are generated by a dedicated \textit{generator} model and validated by a \textit{validator} model before being presented to the tested model, in order to ensure variant quality (98.8\% validator pass rate across configurations).

\noindent\textbf{Models.} We test six distinct configurations (Table~\ref{taba:factorial}) that combine two \textit{tested models} (\gpt, \claude), two \textit{generators} (GPT-5.4, Claude Sonnet 4.5), and two \textit{validators} (GPT-5.2, Claude Haiku 4.5). We assign the stronger model from each family to the generator role and the smaller model to the validator role, motivated by evidence that verification is a comparatively easier capability than generation for language models: models frequently produce correct outputs that weaker models can still validate reliably, even when those same weaker models could not have generated the output themselves \citep{saadfalcon2025weaver}. The factorial structure allows us to assess whether CBI patterns depend on the tested model alone or whether they interact with the perturbation-generation and validation pipeline.

\noindent\textbf{Metrics and definitions.} For each example (a base item $\times$ dimension pair), we record the following. \textit{Consistency} is an entropy-based measure of agreement across the baseline response and all variant responses, with range $[0,1]$, where 1 indicates identical answers across all variants. \textit{Accuracy} is a binary indicator of whether the model answer matches the gold label. Combining both, we define an example as CBI as
\[
\text{CBI}(x) = [\text{consistency}(x) \geq t] \wedge [\text{model\_answer}(x) \neq \text{gold}(x)],
\]

with a default threshold of $t = 0.80$. In addition, we measure \textit{answer collapse}, i.e.\ whether all variant responses produce the identical (wrong) label. This is a stricter condition that indicates deterministic rather than stochastic error.

The joint consistency--accuracy space forms four quadrants (Figure~\ref{fig:quadrant}): Q1 (high consistency, correct) corresponds to reliable behavior; Q2 (low consistency, correct) corresponds to accurate but inconsistent behavior; Q3 (low consistency, incorrect) corresponds to behavior that is unreliable but detectable; and Q4 (high consistency, incorrect) corresponds to CBI, which is not detectable by agreement-based signals.
\vspace{-0.5em}

\section{Results and Discussion}
\vspace{-0.5em}
We organize our findings around five questions, each answered directly from the experimental setup it required.

\begin{table}[t]
\centering
\small
\caption{Aggregated quadrants summary (consistency split t = 0.8).}
\label{tab:aggregated-quadrants}
\begin{tabular}{lrrr}
\toprule
\textbf{Quadrant} & \textbf{N} & \textbf{\% of all observations} & \textbf{\% of incorrect-only} \\
\midrule
Q1 & 19,579 & 84.50 & 0.00 \\
Q2 & 1,577  & 6.81  & 0.00 \\
Q3 & 801    & 3.46  & 39.77 \\
Q4 (CBI) & 1,213  & 5.24  & \textbf{60.23} \\
\bottomrule
\end{tabular}
\label{tab:quadrants}
\end{table}

\begin{figure}[t]
    \centering
    \includegraphics[width=0.7\linewidth]{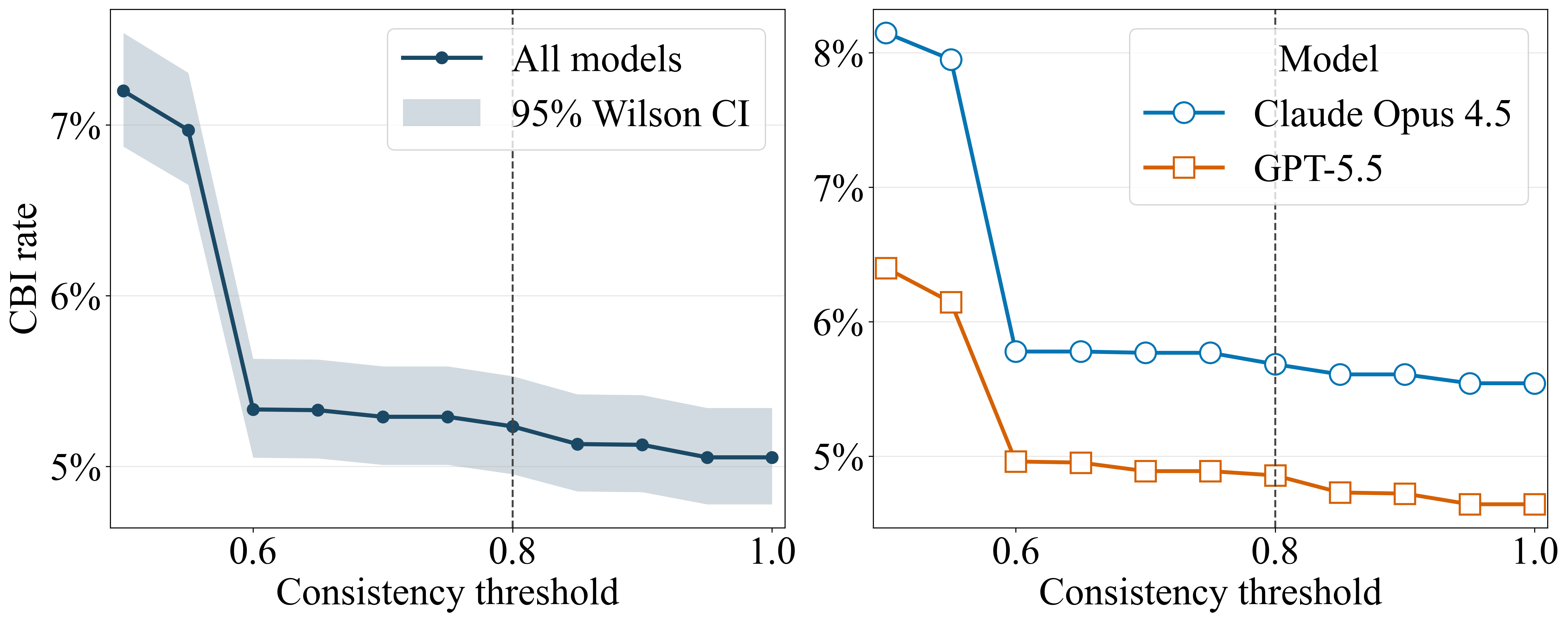}
    \caption{CBI sensitivity, both overall (left) and across models (right).}
    \label{fig:cbi-sens}
\end{figure}

\textbf{How large is the dissociation between consistency and accuracy, and where does the error mass concentrate?} At the standard operating threshold for agreement-based filters (t = 0.80), 5.24\% of all 23,170 observations fall in the high-consistency/incorrect quadrant (Q4, Consistent But Inaccurate, Table~\ref{tab:quadrants}). This rate understates the practical problem: conditioned on being incorrect, 60\% of all errors (1,213 of 2,014) fall in Q4, so an error is about 2.03x more likely to appear in Q4 than in the low-consistency/incorrect quadrant Q3 (1,213 vs. 801). The association between consistency stratum and correctness is highly non-random ($\chi^2(1,N=23,170) = 2085.38, p < 0.001$). Using odds ratio as effect size for the 2x2 table, the odds of being incorrect in high-consistency vs. low-consistency strata are $OR = 0.122$ , equivalently the low-consistency stratum has about 8.20x higher odds of error. Among CBI examples, 77.7\% exhibit complete answer collapse — identical wrong answers across all three perturbation variants — indicating that the dissociation is not a marginal disagreement a majority vote could resolve, but a deterministic convergence on a single incorrect answer.

\textbf{Is CBI an artifact of the consistency threshold, or a structural property of the model's error distribution?} We recompute CBI rates across thresholds from t = 0.50 to t = 1.00. The overall rate moves from 7.2\% to 5.05\%, a compression of only 2.15\% across the full range, and the curve flattens at t = 0.6 (Figure~\ref{fig:cbi-sens}). This indicates that CBI examples are not borderline cases pushed just over an arbitrary cutoff; the bulk sit at or near perfect consistency regardless of where the threshold is drawn. The pattern holds separately for each model family, with Claude Opus 4.5 showing a consistently higher CBI rate (5.54–8.15\%) than GPT-5.5 (4.64-6.4\%) at every threshold tested.

\textbf{Does CBI depend on the tested model, or on the perturbation-generation pipeline?} Across the 6 generator-validator configurations, the overall CBI rate ranges narrowly (from $4.2\%$ to $6.3\%$), and switching validator family (GPT-5.2 vs. Claude Haiku 4.5) shifts CBI rates by 0.57\% for \claude and 0.7\% for \gpt, which indicates the phenomenon is a property of the tested model's learned preferences rather than an artifact of how perturbations were generated or validated. To test whether cross-model overlap could be explained by independent failures, we evaluated the common item set ($N=4{,}264$), where the independence baseline $P_{\text{indep}}(\text{shared error}) = 0.01$ implies an expected shared-error count of $E_{\text{indep}} = 44.92$. Empirically we observe $159$ shared CBI items (and $282$ shared wrong items overall), i.e.\ $3.54\times$ the independence expectation, and a goodness-of-fit chi-square test strongly rejects the independence account, $\chi^2(1)=292.76$, $p=1.24\times 10^{-65}$. Because these failures are not independent, cross-model ensembling — the strongest available mitigation against single-model unreliability — does not detect them.

\begin{table}[t]
\centering
\caption{Dataset-level CBI rates and counts.}
\label{tab:rq4}
\begin{tabular}{lccccc}
\toprule
\textbf{Dataset} & \textbf{CBI \% overall} & \textbf{CBI \% \gpt} & \textbf{CBI \% \claude} & \textbf{N errors} & \textbf{N CBI} \\
\midrule
\cfa & 3.80 & 2.46 & 5.21  & 515 & 248 \\
\cfat & 5.22 & 4.77 & 5.99 & 542 & 317 \\
\cmt       & 5.42 & 5.26 & 5.56  & 630 & 377 \\
\fq     & 7.49 & 8.22 & 6.51  & 327 & 271 \\
\bottomrule
\end{tabular}
\end{table}

\begin{figure}[t]
    \centering
    \includegraphics[width=0.7\linewidth]{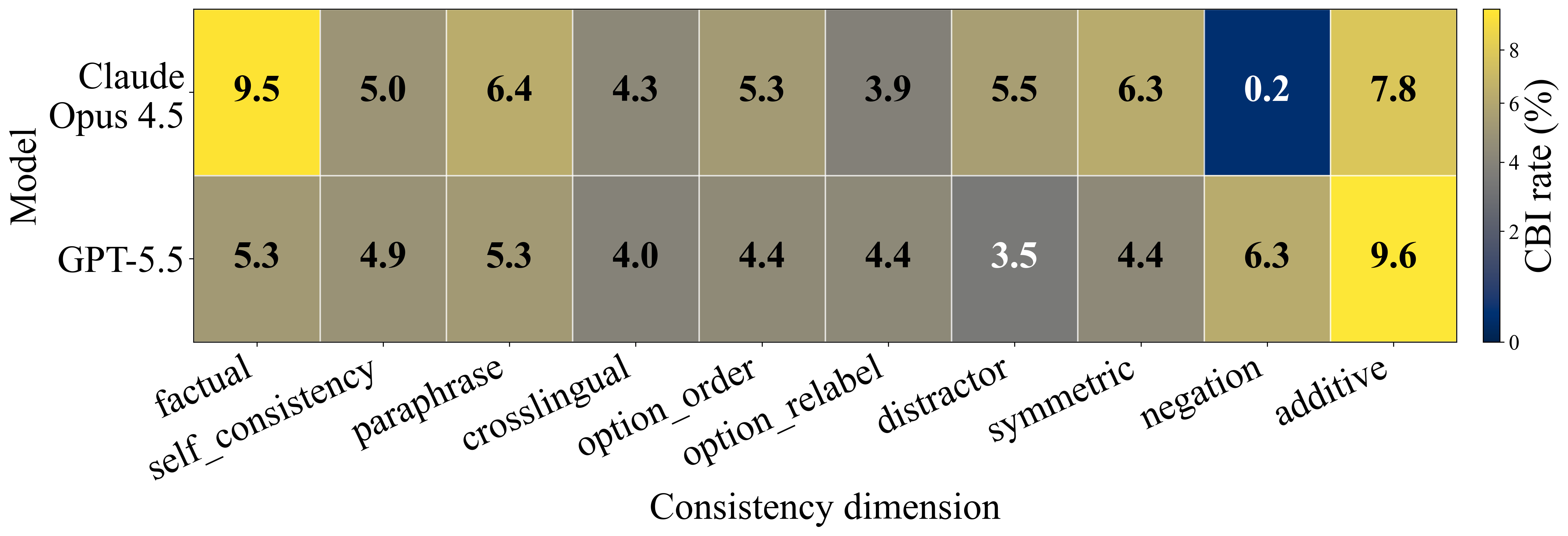}
    \caption{CBI rate (\%) by model and consistency dimension.}
    \label{fig:cbi-cons-cat}
\end{figure}


\textbf{Where does CBI concentrate across domain difficulty, question format, and perturbation type?} CBI rate increases with question complexity: 5.22\% for \cfat (multi-step quantitative reasoning) versus 3.80\% for \cfa (Table~\ref{tab:rq4}). Comparing formats, \cmt (multiple-choice) shows a lower rate (5.42\%) than \fq (free-form numerical reasoning, 7.49\%), and this gap does not depend on the model. Breaking down by perturbation dimension (Table~\ref{taba:heatmap}), CBI is broadly distributed rather than concentrated in any single dimension: rates fall between roughly 3.5\% and 9.7\% for nine of the ten dimensions and for both models, with additive perturbation the highest for each (9.65\% for \gpt, 7.83\% for \claude) and factual perturbation the next highest for \claude (9.48\%). Aggregate CBI rates across dimensions are nearly identical for the two models (5.42\% and 5.20\%), but the dimension-level profiles differ. 

\textbf{Is CBI stochastic noise, or evidence of a stable, mechanistic failure?} The 77.7\% answer-collapse rate among CBI examples (RQ1) is the central piece of evidence here: a model producing the identical wrong answer across meaning-preserving perturbation is not exhibiting noisy disagreement that would wash out under resampling, but a stable, reproducible preference. The dimension-level breakdown (Table~\ref{taba:ans-collapse}) shows that this property is not driven by any single dimension. Collapse is complete (100\%) for option order, option relabel, and distractor rewording, remains high for symmetric (93.0\%), negation (85.7\%), and crosslingual (79.5\%) perturbation, and even at the low end retains roughly two-thirds for factual (74.8\%), self-consistency (69.7\%), and paraphrase (68.5\%), so that for most CBI examples the model does not merely agree with itself at the level of a consistency score but returns the same specific wrong answer. That collapse is strongest precisely on the surface-form dimensions agreement-based filters are designed to probe is where a consistency signal would be most confidently and most wrongly reassuring. Negation warrants separate interpretation, since its underlying failure may be an inability to detect semantic inversion rather than the trained-in wrong-preference mechanism driving the other dimensions, but the two are operationally indistinguishable to any consistency-based filter, which is the property that matters for downstream pipeline design.
\vspace{-0.5em}
\subsection{Limitations}
\vspace{-0.5em}
Our findings are demonstrated within financial QA. Whether the same quadrant mass and the same concentration patterns hold in other specialized domains, such as legal, biomedical, or engineering QA, remains an empirical question, although the cross-model convergence we observe suggests that the mechanism is not specific to finance.

\vspace{-0.5em}
\subsection{Implications for Evaluation Pipelines}
\vspace{-0.5em}
Our results do not show that consistency is uninformative; the concern is rather with
what a consistency-based filter leaves behind. Such a filter certifies 89.7\% of all
items, and because almost all items fall in the high-consistency stratum, 60\% of the
total error mass ends up inside the certified set, so that gating buys approximately a
one-third reduction in error rate while leaving most errors in the region it has labeled
reliable. What is consequential about this residual is its character rather than its
size: among CBI examples, 77.7\% show complete answer collapse, returning the same
specific wrong answer across every perturbation variant, so these are not borderline
cases sitting just above the threshold but deterministic preferences against which the
two standard responses to unreliable output, raising the threshold and cross-model
ensembling, are both ineffective.

\paragraph{No single dimension to exclude.} CBI is distributed across dimensions rather
than concentrated in any one of them, which has a direct practical consequence: if a
single perturbation type drove the dissociation, a pipeline could exclude that dimension
and recover a usable signal, but our results indicate that no such dimension exists.
Answer collapse is moreover highest on precisely those surface-form dimensions that
agreement-based filters are designed to probe, reaching 100\% for option order, option
relabel, and distractor rewording, which is where a consistency signal is at once most
reassuring and most misleading.

\paragraph{Bearing on the inter-judge agreement question.} Agreement and correctness are
related, but the relationship is conditional on domain and dimension, so the quantity
that determines deployment risk is not a single transportable threshold but the error
mass that remains above it. Because our perturbations preserve meaning by construction,
the failures we observe are categorically different from formatting sensitivity: the
model is not reacting to surface form, but is stably wrong regardless of it, which is a
harder case for any surface-sensitivity mitigation to catch.

\paragraph{Practical recommendations.} We identify two concrete uses of quadrant analysis
in the design of evaluation pipelines. First, as a \textit{diagnostic before deployment},
the mass of Q4 should be estimated on a small labeled sample before a consistency-based
filter is trusted in a new domain, rather than assuming that general-domain calibration
transfers. Second, as a form of \textit{structured disclosure}, the error mass remaining
in the certified set should be reported alongside aggregate consistency, since it is this
quantity that determines whether an agreement-based signal is adequate for a given
deployment. We propose three fields for evaluation disclosure templates: the
certified-set error rate, the CBI-within-inaccurate rate, and the answer-collapse rate,
stratified by perturbation type.

\vspace{-0.5em}
\section{Conclusion}
\vspace{-0.5em}
Consistency carries real signal about correctness in financial QA, but it is not a
substitute for it. Because almost all items fall in the high-consistency region, a
standard agreement-based filter certifies 90\% of our data while keeping 60\% of its
errors, and 77.7\% of those retained errors are identical wrong answers repeated
across every perturbation variant. Neither a stricter threshold nor cross-model
ensembling removes them, since the rate is nearly flat across thresholds and the
failures are shared across model families. Consistency-based gating should therefore
be treated as an error-rate reduction with a largely irreducible residual rather than
a stand-in for ground truth, and we suggest that this residual mass become a standard
part of evaluation pipeline disclosure.

{
\small

\bibliography{refs}
\bibliographystyle{apalike}
}







\newpage
\appendix
\crefalias{section}{appendix}
\section*{Appendix}

\setcounter{figure}{0}
\renewcommand{\thefigure}{A.\arabic{figure}}

\setcounter{table}{0}
\renewcommand{\thetable}{A.\arabic{table}}

\section{Details on Experimental Setup}

\begin{table}[htbp]
\centering
\caption{Configurations of model setup}
\begin{tabular}{lll}
\toprule
Tested models & Generators & Validators \\
\midrule
GPT 5.5 & GPT 5.4 & GPT 5.2 \\
GPT 5.5 & Claude Sonnet 4.5 & Claude Haiku 4.5 \\
GPT 5.5 & GPT 5.4 & Claude Haiku 4.5 \\
Claude Opus 4.5 & GPT 5.4 & GPT 5.2 \\
Claude Opus 4.5 & Claude Sonnet 4.5 & GPT 5.2 \\
Claude Opus 4.5 & Claude Sonnet 4.5 & Claude Haiku 4.5 \\
\bottomrule
\end{tabular}
\label{taba:factorial}
\end{table}

\subsection{Scoring}

\textbf{Consistency score.} Within-set agreement of the answer set (baseline + variants), computed in the canonical frame. The default MCQ formula is normalized Shannon entropy: $1 - H / log(n)$ where n is the set size (all-agree $\to$ 1.0, all-distinct $\to$ 0.0). Unparsed answers count against consistency; they never agree with anything.

\section{Extended CBI Breakdown by Dimensions and Models}

\begin{table}[htbp]
\centering
\caption{CBI rate by model and consistency dimension}
\begin{tabular}{lll}
\toprule
model & Claude Opus 4.5 & GPT-5.5 \\
\midrule
factual & 9.48 & 5.28 \\
self\_consistency & 4.98 & 4.86 \\
paraphrase & 6.38 & 5.27 \\
crosslingual & 4.31 & 3.98 \\
option\_order & 5.32 & 4.42 \\
option\_relabel & 3.9 & 4.35 \\
distractor & 5.52 & 3.49 \\
symmetric & 6.27 & 4.41 \\
negation & 0.24 & 6.26 \\
additive & 7.83 & 9.65 \\
\bottomrule
\end{tabular}
\label{taba:heatmap}
\end{table}

\begin{table}[htbp]
\centering
\caption{Share of CBI rows with identical samples answers by consistency dimension}
\begin{tabular}{llll}
\toprule
Dimension & N CBI & N identical answers & Identical share \% \\
\midrule
option\_order & 136 & 136 & 100.0 \\
option\_relabel & 116 & 116 & 100.0 \\
distractor & 80 & 80 & 100.0 \\
symmetric & 57 & 53 & 92.98 \\
negation & 42 & 36 & 85.71 \\
crosslingual & 117 & 93 & 79.49 \\
factual & 258 & 193 & 74.81 \\
self\_consistency & 175 & 122 & 69.71 \\
paraphrase & 165 & 113 & 68.48 \\
additive & 67 & 0 & 0.0 \\
\bottomrule
\end{tabular}
\label{taba:ans-collapse}
\end{table}

\end{document}